\documentclass{article}
\usepackage{spconf,amsmath,graphicx}
\usepackage{algorithm}
\usepackage{textcomp}
\usepackage{xcolor}
\usepackage{subcaption}
\usepackage{amsmath}

\usepackage{amsmath, amssymb}
\usepackage{amsmath}
\usepackage{amsfonts}
\usepackage{algpseudocode} 

\title{Successive Interference Cancellation-aided Diffusion Models for Joint Channel Estimation and Data Detection in Low Rank Channel Scenarios}
\name{Sagnik Bhattacharya, Muhammad Ahmed Mohsin, Kamyar Rajabalifardi, John M. Cioffi\thanks{Emails: \{sagnikb, muahmed, kfardi, cioffi\}@stanford.edu}}
\address{Dept. of Electrical Engineering, Stanford University, Stanford, CA, USA}

\begin{document}
%
\maketitle
\begin{abstract}
This paper proposes a novel joint channel-estimation and source-detection algorithm using successive interference cancellation (SIC)-aided generative score-based diffusion models. Prior work in this area focuses on massive MIMO scenarios, which are typically characterized by full-rank channels, and fail in low-rank channel scenarios. The proposed algorithm outperforms existing methods in joint source-channel estimation, especially in low-rank scenarios where the number of users exceeds the number of antennas at the access point (AP). The proposed score-based iterative diffusion process estimates the gradient of the prior distribution on partial channels, and recursively updates the estimated channel parts as well as the source. Extensive simulation results show that the proposed method outperforms the baseline methods in terms of normalized mean squared error (NMSE) and symbol error rate (SER) in both full-rank and low-rank channel scenarios, while having a more dominant effect in the latter, at various signal-to-noise ratios (SNR).
\end{abstract}
\begin{keywords}
Joint Langevin, Diffusion Model, SIC, Channel Estimation, Score-based Generative Model, Data Detection.
\end{keywords}
\section{Introduction}
\label{sec:intro}

Massive multiple-input multiple-output (mMIMO) systems are vital for next-generation wireless networks, especially in environments demanding better spatial resolution ~\cite{10422885}. These systems use numerous base station antennas for beamforming, enabling high-gain, narrow beams for multi-user communication ~\cite{song2022benchmarking}. However, massive MIMO assumes a full-rank wireless channel, valid in the uplink only if the number of antennas at the AP or BS equals or exceeds the number of users and sufficient scattering is present. Research has explored joint channel estimation and symbol detection in massive MIMO ~\cite{nair2022joint,kong2020reducing,e25091358}, with least square (LS) estimators being optimal for rich multipath channels ~\cite{9252921}. As the number of users in wireless applications grows, low-rank channels become a challenge, where users exceed the number of antennas at the AP/BS. The minimum mean squared error (MMSE) estimator may struggle in THz and mmWave systems due to beamspace sparsity in high-dimensional channels ~\cite{8732419}. MIMO detection is NP-hard ~\cite{delpia2014mixedintegerquadraticprogrammingnp}, and with $N_u$ users and $M$-ary modulation, maximum likelihood (ML) decoding complexity scales exponentially as $\mathcal{O}(M^{N_u})$, making it impractical for large systems. For SIMO systems, gradient descent-based methods have been proposed ~\cite{castañeda2017vlsidesignsjointchannel}, while in MIMO, an interference cancellation information geometry approach ~\cite{lu2024interferencecancellationinformationgeometry} achieves near-MMSE performance with reduced complexity.


Diffusion models improve MIMO channel estimation by learning complex data distributions and leveraging prior information for accurate estimation. They model the score function of the likelihood function using variational lower bounds, enhancing generative performance~\cite{wu2023cddmchanneldenoisingdiffusion}. For example, ~\cite{fesl2024diffusionbasedgenerativepriorlowcomplexity} employs a diffusion model with a compact CNN architecture, utilizing channel sparsity in the angular domain for simplified estimation. In ~\cite{rice}, a score-based diffusion model is used for joint channel estimation and data detection in multi-user MIMO systems through an iterative process.

Despite their success, diffusion-based channel estimation methods struggle in low-rank channel scenarios. This paper presents a novel algorithm for joint channel estimation and data detection in such scenarios using score-based diffusion models. The proposed method employs a successive interference cancellation (SIC)-aided diffusion model to estimate gradients for prior distributions of partial channels, which are iteratively updated along with detected data symbols and the dynamic SIC decoding order.\\
\textbf{Contributions:} The contributions of this paper are:\\
\textbf{1.} A novel SIC-aided joint channel estimation and data detection algorithm using score-based diffusion models for partial channel prior distribution estimation.  \\
\textbf{2.} A channel gain-based SIC decoding order which is iteratively updated, and determines the partial channels to be estimated. \\
\textbf{3.} Extensive simulation evaluation demonstrating the proposed algorithm outperforming baseline algorithms across low-rank and full-rank channels, for different hyperparameter settings.

\section{Problem Formulation}
\label{sec: prob_form}

We examine an uplink multiple access channel (MAC) where $N_u$ single-antenna transmitters send pilot and data symbols to an access point (AP) equipped with $N_r$ antennas. Our focus is on the low-rank channel scenario where the number of users exceeds the number of antennas at AP, i.e. $N_u \geq N_r$. In this system, specific time slots are allocated for transmitting pilot signals and data. Using this time-division approach, the mathematical model for the uplink MAC communication system is given by:
\begin{equation} \mathbf{Y} = \mathbf{H} \mathbf{X} + \mathbf{Z}, \end{equation}
where $\mathbf{H} \in \mathbb{C}^{N_r \times N_u}$ represents the channel matrix. The noise matrix $\mathbf{Z}$ consists of columns $Z_j$ for $j=1,…,K$, each independently drawn from a zero-mean complex Gaussian distribution with covariance matrix $\sigma_0^2 \mathbf{I}_{N_r}$. $\mathbf{X} = [\mathbf{X_P}, \mathbf{X_D}] \in \mathcal{X}^{N_u \times K}$ combines both the pilot and data sequences, where $\mathbf{X_P} \in \mathcal{X}^{N_u \times P}$ represents the pilot sequence, transmitted during the first $P$ time slots and $\mathbf{X_D} \in \mathcal{X}^{N_u \times D}$ represents the unknown data sequence transmitted at $D$ time slots ($K = P + D$). The set $\mathcal{X}$ denotes a finite constellation of possible symbol values from quadrature amplitude modulation (QAM). We assume that all symbols are transmitted with equal power. Also, $\mathbf{Y} \in \mathbb{C}^{N_r \times 
K}$ are the observations at the receiver. It is further assumed that the receiver does not have prior knowledge of the channel matrix $\mathbf{H}$, although the noise power $\sigma_0^2$ is known. Consequently, we utilize a Maximum A Posteriori (MAP) detector to jointly estimate $\mathbf{X}_{\mathbf{D}}$ and $\mathbf{H}$ based on the known pilot sequence $\mathbf{X}_{\mathbf{P}}$ and the received data $\mathbf{Y}$. In this low-rank channel context, we employ SIC to sequentially decode each user's signal \cite{saito2013non}, \cite{islam2016power}, \cite{wang2023fairness}.

\section{SIC-aided Langevin Diffusion for Low-rank Channel}
\label{sec: algorithm}
In our framework, we jointly estimate the channel and detect symbols in low-rank uplink channel scenarios, i.e., when the number of users exceeds the number of AP/BS antennas. The proposed algorithm uses SIC-aided scoring diffusion for channel prior estimation, based on Langevin dynamics \cite{pavliotis2014stochastic}. For symbol detection, given the estimated channel, we use SIC. We define $\boldsymbol{\pi}(.)$ as the SIC decoding order vector such that $\boldsymbol{\pi}(i)$ is the decoding order for user $i$, and $\boldsymbol{\pi}^{-1}(.)$ as its inverse, where $\boldsymbol{\pi}^{-1}(i)$ indicates the user index decoded at $i^{\textrm{th}}$ stage in SIC. Moreover, we denote $\boldsymbol{\pi}^{-1}(i:j)$ as the set of users decoded from $i^{\textrm{th}}$ stage to $j^{\textrm{th}}$ stage. We split the channel matrix $\mathbf{H}$ into $\lceil \frac{N_u}{N_r}\rceil$ smaller submatrices of size $N_r \times N_r$ each; where we define $\mathbf{H}^{(i)} = \mathbf{H}(\boldsymbol{\pi}^{-1}(iN_r : \min\{(i+1)N_r - 1, N_u\}))$ as the channel submatrix corresponding to users at decoding order indices $iN_r$ to $\min\{(i+1)N_r - 1, N_u\}$. Correspondingly, we define  $\mathbf{X}_{\mathbf{D}}^{(i)} = \mathbf{X}_{\mathbf{D}}(\boldsymbol{\pi}^{-1}(iN_r : \min\{(i+1)N_r - 1, N_u\}))$ and $\mathbf{X}_{\mathbf{P}}^{(i)} = \mathbf{X}_{\mathbf{P}}(\boldsymbol{\pi}^{-1}(iN_r : \min\{(i+1)N_r - 1, N_u\}))$ as the $i^{th}$ partial data symbol and pilot symbol sets respectively, which pass through the channel submatrix $\mathbf{H}^{(i)}$.

The scoring diffusion model estimates the prior distribution for each of these partial channel submatrices individually. We do this to make each of the individual submatrices close to full-rank (to the extent that the environment is rich enough ans the noise added by the other channel submatrix components). This emulates the massive MIMO scenario for every submatrix, and scoring diffusion models perform well on approximately full-rank channels, as shown in \cite{rice}. However, a crucial difference from massive MIMO in our approach is that the output $\mathbf{Y}$ here is a linear combination of partial inputs sent through the partial channel submatrices. This can be expressed as 
$\mathbf{Y} = \mathbf{H}^{(1)}\mathbf{X}^{(1)} + \mathbf{H}^{(2)}\mathbf{X}^{(2)} + \cdots + \mathbf{H}^{(k)}\mathbf{X}^{(k)} + \mathbf{Z}$ where $k = \lceil\frac{N_u}{N_r}\rceil$, and $\mathbf{X}^{(i)} = [\mathbf{X}_{\mathbf{P}}^{(i)}, \mathbf{X}_{\mathbf{D}}^{(i)}]$.
Hence, unlike massive MIMO receiver, in this case, to detect submatrix $\mathbf{H}^{(1)}$ and partial symbols $\mathbf{X}^{(1)}$ with output $\mathbf{Y}$ given, we have to eliminate the components of $\mathbf{Y}$ coming from the other submatrices, as well as the noise.

Taking into account our split channel submatrices, we formulate the joint channel estimation and symbol detection using MAP estimation as:
\begin{equation} \label{eq:MAP}
    \begin{aligned}
        \mathcal{L} =& \log {\left(p\left(\mathbf{X}_{\mathbf{D}}, \mathbf{H} \mid \mathbf{Y}, \mathbf{X}_{\mathbf{P}}\right)\right)} \\
         = &\sum_{i=1}^{\lceil \frac{N_u}{N_r}\rceil}\log{\left(p\left(\mathbf{Y}\mid \mathbf{X}^{(i-1)}, \cdots \mathbf{X}^{(1)}, \mathbf{H}^{(i-1)}, \cdots \mathbf{H}^{(1)}\right) \right)} \\
        & \quad\quad\:+ \operatorname{log}\left(p\left(\mathbf{X}^{(i)}\right)\right) + \operatorname{log}\left(p\left(\mathbf{H}^{(i)}\right)\right) \\
    \end{aligned}
\end{equation}
Using Bayes' rule, this boils down to

\begin{equation}
    \begin{aligned}
     \mathcal{L}   = & \sum_{i=1}^{\lceil \frac{N_u}{N_r}\rceil}\log{\left(\mathcal{N}\left(\mathbf{H}^{(i-1)}\mathbf{X}^{(i-1)} + \cdots + \mathbf{H}^{(1)}\mathbf{X}^{(1)}, \Sigma_i\right)\right)} \\
    & \quad\quad\:+ \operatorname{log}\left(p\left(\mathbf{X}^{(i)}\right)\right) + \operatorname{log}\left(p\left(\mathbf{H}^{(i)}\right)\right)       
    \end{aligned}
\end{equation}
In the summation above, the first term corresponds to a normal distribution according to the previous symbols detected. $\Sigma_i$ is the covariance matrix corresponding to the SIC decoding order, i.e. considering the users which appear later in the decoding order as noise, i.e. $\Sigma_i = \mathbf{Z} + \sum_{i+1}^{U}\mathbf{H}^{(i)}\mathbf{X}^{(i)}$. 
\begin{figure*}[t!]
    \centering
    \begin{minipage}[b]{0.45\linewidth}
        \centering
        \includegraphics[width=0.8\linewidth]{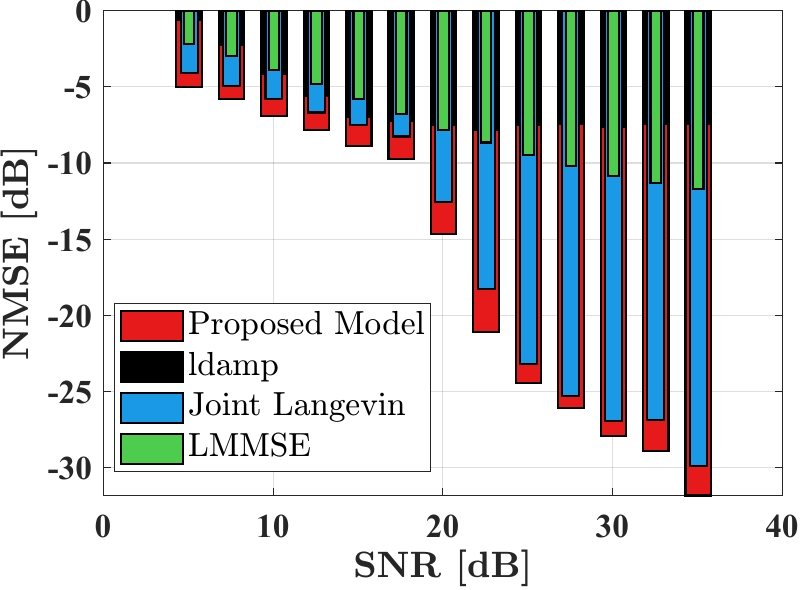}
        \subcaption{ NMSE for high rank}
        \label{fig:nmse_highrank}
    \end{minipage}
    \hfill
    \begin{minipage}[b]{0.45\linewidth}
        \centering
        \includegraphics[width=0.8\linewidth]{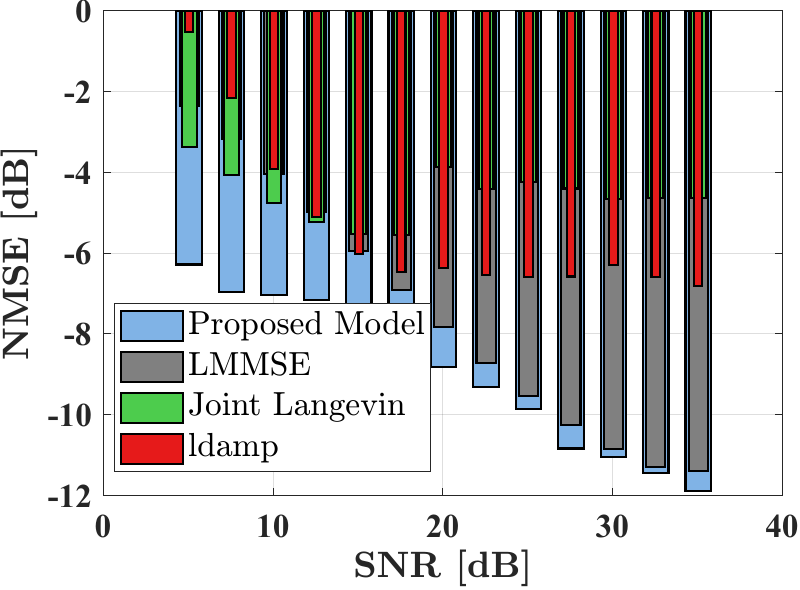}
        \subcaption{NMSE for low rank}
        \label{fig:nmse_lowrank}
    \end{minipage}

    \vspace{0.3cm} 

    \begin{minipage}[b]{0.45\linewidth}
        \centering
        \includegraphics[width=0.8\linewidth]{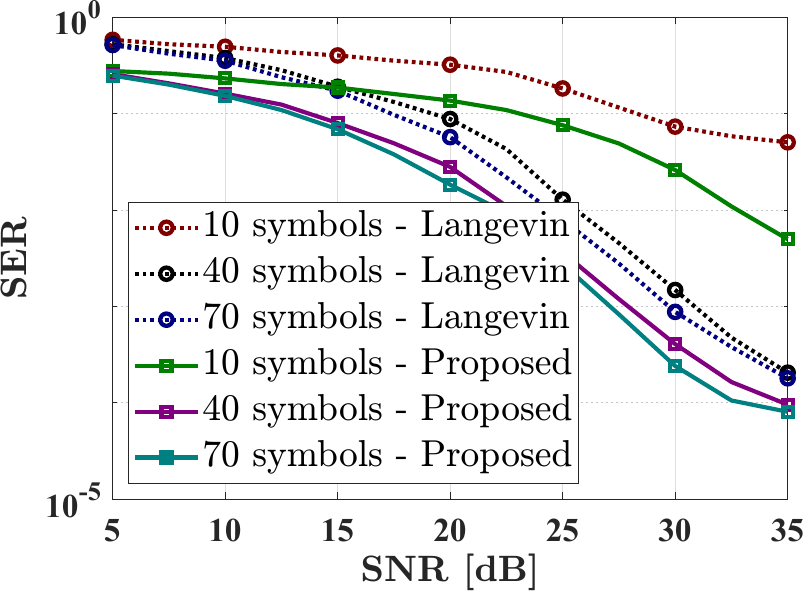}
        \subcaption{ SER for high rank}
        \label{fig:ser_highrank}
    \end{minipage}
    \hfill
    \begin{minipage}[b]{0.45\linewidth}
        \centering
        \includegraphics[width=0.8\linewidth]{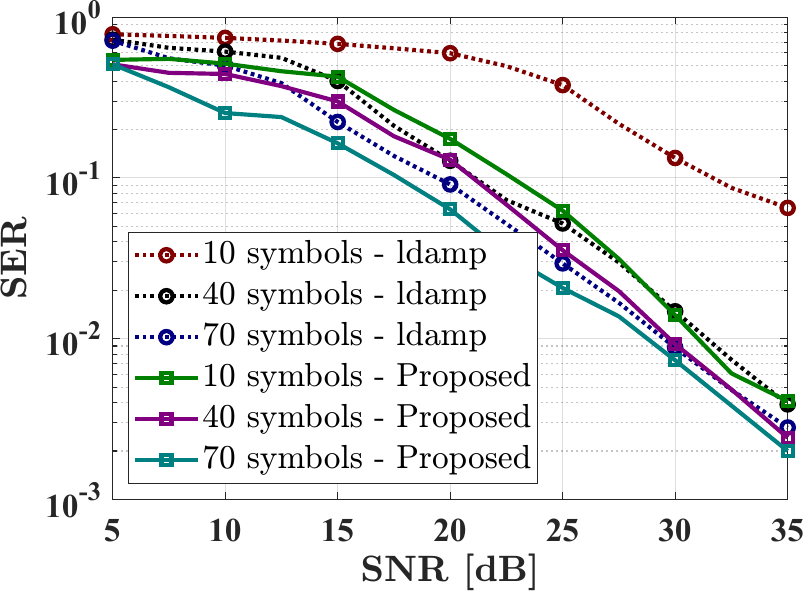}
        \subcaption{ SER for low rank}
        \label{fig:ser_lowrank}
    \end{minipage}
    
    \caption{Comparison of our proposed method with different baselines, we use 3GPP channel models with number of pilots $P = 30$, and the number of symbols $D = 50$}
    \label{fig:overall}
\end{figure*}

The second and third terms are the prior probability density functions of symbols transmitted and the channel matrix. To maximize MAP function $\mathcal{L}$, we adopt an iterative approach and update variables $\mathbf{X}^{(i)}$, and $\mathbf{H}^{(i)}$ alternately. For the gradient of the prior of $\mathbf{H}^{(i)}$, we train a score network $s_{\theta}(\mathbf{H}^{(i)}, \sigma_{i, \mathbf{H}})$. This score network is a diffusion model based on Langevin dynamics. For maximization, we now compute the gradient of $\mathcal{L}$ with respect to the variables $\mathbf{H}^{(i)}$ and $\mathbf{X}^{(i)}$. Therefore, the gradient of $\mathcal{L}$ with respect to $\mathbf{H}^{(i)}$ can be expressed as:
\begin{equation}
\begin{aligned}
      \nabla_{\mathbf{H}^{(i)}} \mathcal{L} = &\sum_{j=i}^{\lceil \frac{N_u}{N_r}\rceil}  \Sigma_j^{-1} \left(\mathbf{Y}-{\mathbf{H}}^{(j)}{\mathbf{X}}^{(j)}\right)\left[{\mathbf{X}}_{\mathbf{D}}^{(j)}, \mathbf{X}_{\mathbf{P}}\right]^{\mathrm{*}} \\
      & + s_{\theta} (\mathbf{H}^{(i)})
\end{aligned}
\end{equation}

\begin{equation}\label{eq:gradX}
    \begin{aligned}
        \nabla_{\mathbf{X}_{\mathbf{D}}^{(i)}} \mathcal{L} = &\sum_{j=i}^{\lceil\frac{N_u}{N_r}\rceil}{\mathbf{H}^{(j)}}^*\Sigma_j^{-1}\left(\mathbf{Y}_{\mathbf{D}}-\mathbf{H}^{(j)} \mathbf{X}_{\mathbf{D}}^{(j)}\right) \\
        & + \nabla_{\mathbf{X}_{\mathbf{D}}^{(i)}} \log{p(\mathbf{X}_{\mathbf{D}}^{(i)})}
    \end{aligned}
\end{equation}
In equation (\ref{eq:gradX}), the term $\nabla_{\mathbf{X}_{\mathbf{D}}^{(i)}} \log{p(\mathbf{X}_{\mathbf{D}}^{(i)})}$ is approximated by a continuous approximation of the discrete input distribution by applying an annealing process with variable noise levels, adopted from \cite{rice}. 
To summarize the entire algorithm, we iteratively alternate between SIC decoding order derivation, partial channel estimation, corresponding partial symbol detection, until convergence is reached. The SIC decoding order $\pi$ at any given iteration is determined as the order of the current channel strengths of the users, i.e., the user with the highest norm of the channel is decoded first, followed by the second highest one, and so on.






\section{Performance Evaluation}
\label{sec: results}
In this section, we evaluate the efficacy of the proposed algorithm in Section~\ref{sec: algorithm} against baseline algorithms~\cite{metzler2017learneddampprincipledneural, rice, 8100963}. \cite{metzler2017learneddampprincipledneural} uses an iterative method that integrates a learned denoising neural network (DnCNN) into the Approximate Message Passing framework to efficiently recover signals from noisy measurements. \cite{rice} uses the Langevin diffusion-based joint channel estimation and data detection; however, they use this for massive MIMO setting and do not employ partial channel submatrix prior estimation and dynamic SIC. Finally, the third baseline \cite{8100963} uses linear minimum mean squared error (LMMSE) minimization to estimate the channel and data symbols. The first experiment benchmarks the efficacy of our approach for low-rank channel scenarios with baseline methods~\cite{metzler2017learneddampprincipledneural, rice, assalini2010robustness} for channel estimation. The second experiment demonstrates the benchmarks of our proposed methodology against the same baseline methods for low-rank channel scenarios. The third experiment investigates the impact of varying the number of transmitted data symbols ($\mathbf{X_D}$) on the symbol error rate (SER) as a function of signal-to-noise ratio (SNR), while keeping the number of pilot symbols constant. This analysis explores how the system's performance changes when different amounts of data are transmitted alongside a fixed pilot sequence used for channel estimation, and compares the results with \cite{rice} for high-rank channel scenarios. The final result demonstrates the SER against SNR for low-rank channel scenarios and compares it with the L-DAMP model. \\
\textbf{Signal modeling and simulation:} For our experiments, we utilize QuaDRiGa \cite{jaeckel2014quadriga} to generate 3GPP 3D channel models. Specifically, we generate 10,000 instances of low-rank channels with $N_u = 64$ and $N_r = 32$, following the configurations detailed in ~\cite{fesl2024diffusionbasedgenerativepriorlowcomplexity}.\\
\textbf{Channel estimation for high-rank scenarios:} In Fig.~\ref{fig:nmse_highrank}, we demonstrate the performance analysis for our proposed methodology against similar baselines methods for high rank channel scenarios. Again, for the simulation, we keep $P = 30$ and $D = 50$ and observe that our proposed method consistently outperforms the baseline methods for all SNRs.\\
\textbf{Channel estimation for low-rank scenarios:} In Fig.~\ref{fig:nmse_lowrank}, we demonstrate the NMSE vs. SNR for low-rank channel scenarios. We compare our results with baseline methods for channel estimation, such as L-DAMP~\cite{metzler2017learneddampprincipledneural}, L-MMSE~\cite{8100963} and Joint Langevin~\cite{rice}. Our methodology outperforms all baseline methods for all SNRs, particularly for lower SNRs, where the proposed methodology exceeds with several orders of magnitude. For the simulation, we set $P = 30$ and $D = 50$.\\
\textbf{SER for high-rank scenarios:} In Fig.~\ref{fig:ser_highrank}, we analyze the SER performance for high-rank channel scenarios. We evaluate our model for $\mathbf{
X_D = [10, 40, 70]}$ against Joint Langevin method. We observe that as $\mathbf{X_D}$ increases, the SER efficacy increases. The improvement provided by the proposed algorithm over the baselines shows significant increase with increasing SNR\\
\textbf{SER for low-rank scenarios:} In Fig.~\ref{fig:ser_lowrank}, we compare the SER for low-rank channels for $\mathbf{X_D = [10, 40, 70]}$ against the L-DAMP method. Our proposed methodology performs better across all SNRs, with performance increasing as SNR increases. For both SER performance graphs, we consider $P = 30$.

\section{Conclusion and Future work}
\label{sec: results}
In this paper, we present a novel SIC-aided scoring diffusion-based joint channel estimation and data detection pipeline for low-rank channel scenarios. Our proposed methodology is based on channel gain based dynamic decoding order for SIC using diffusion model. Simulations reveal the efficacy of our proposed approach over the baseline methods for both rank-sufficient and rank-deficient channels over all SNRs. For future work, enhancing performance over lower SNRs using reinforcement learning for optimum sub-carrier power allocation can be integrated in the pipeline. Furthermore, diffusion models with hyperparameter tuning and efficient search space exploration can further enhance results.

\bibliographystyle{IEEEbib}
\bibliography{strings,refs}

\end{document}